\documentclass[conference]{IEEEtran}
\usepackage{times}
\pdfoutput=1
\usepackage[numbers]{natbib}
\usepackage{multicol}
\usepackage[bookmarks=true]{hyperref}

\usepackage[inline]{enumitem}
\usepackage[utf8]{inputenc}
\usepackage{graphicx}
\usepackage{amsmath} 
\usepackage{amssymb}
\usepackage{xcolor}
\usepackage{pifont}
\usepackage{multirow}
\usepackage{makecell}
\usepackage{tabularx} 
\usepackage{array} 
\usepackage{listings}
\begin{document}

\title{\textbf{\textit{\textbf{\textit{Multi-RAG}}}}: A Multimodal Retrieval-Augmented Generation System for Adaptive Video Understanding}

\author{
  Mingyang Mao\textsuperscript{1, \dag },
  Mariela M. Perez-Cabarcas\textsuperscript{2, \dag},
  Utteja Kallakuri\textsuperscript{1},
  Nicholas R. Waytowich\textsuperscript{2},\\
  Xiaomin Lin\textsuperscript{1},
  Tinoosh Mohsenin\textsuperscript{1} \\[1ex]
  \textsuperscript{1}Johns Hopkins Whiting School of Engineering, Baltimore, Maryland, USA \\
  Emails: \{mmao4, xlin52, ukallak1, tinoosh\}@jhu.edu; \\
  \textsuperscript{2}DEVCOM Army Research Laboratory, Aberdeen, MD, USA \\
  Emails: \{mariela.m.perez-cabarcas.civ, nicholas.r.waytowich.civ\}@army.mil\\
  \textsuperscript{\dag}~Equal contribution.
}


%

\maketitle

\IEEEaftertitletext{
\vspace{-1.2\baselineskip}
\begin{center}
\footnotesize{}
\end{center}
}

\begin{abstract}
To effectively engage in human society, the ability to adapt, filter information, and make informed decisions in ever-changing situations is critical. As robots and intelligent agents become more integrated into human life, there is a growing opportunity—and need—to offload the cognitive burden on humans to these systems, particularly in dynamic, information-rich scenarios.

To fill this critical need, we present \textbf{\textit{Multi-RAG}}, a multimodal retrieval-augmented generation (RAG) system designed to provide adaptive assistance to humans in information-intensive circumstances. Our system aims to improve situational understanding and reduce cognitive load by integrating and reasoning over multi-source information streams, including video, audio, and text. As an enabling step toward long-term human-robot partnerships, \textbf{\textit{Multi-RAG}} explores how multimodal information understanding can serve as a foundation for adaptive robotic assistance in dynamic, human-centered situations. To evaluate its capability in a realistic human-assistance proxy task, we benchmarked \textbf{\textit{Multi-RAG}} on the MMBench-Video dataset, a challenging multimodal video understanding benchmark. Our system achieves superior performance compared to  existing open-source video large language models (Video-LLMs) and large vision-language models (LVLMs), while utilizing fewer resources and less input data. The results demonstrate \textbf{\textit{Multi-RAG}}’s potential as a practical and efficient foundation for future human-robot adaptive assistance systems in dynamic, real-world contexts.\\


\end{abstract}

\IEEEpeerreviewmaketitle

\section{Introduction}

In today's highly social and dynamic settings, we are often overloaded with information. We actively or passively receive vast amounts of information constantly from various media and scenarios \cite {Dealing_with_information_overload}. For example, students face a significant influx of knowledge from both textbooks and teachers in the classroom. Similarly, Army commanders issue orders according to their extensive military training and how it applies to the highly complex situation at hand. These scenarios are all information-intensive. The inability to cope with excessive information can lead to serious performance losses and potential health problems \cite{Arbeitsunterbrechungen_und_Multitasking, girard2008information}.  Therefore, it is crucial to extract valuable information from the abundance of information streams to enhance efficiency in communication, learning, and decision-making. 


Large Language Models (LLMs) are gaining popularity and influencing all aspects of daily life. LLMs show strong potential in improving information summarization techniques by generating concise, coherent representations of lengthy texts while retaining essential content, offering significant utility across a variety of domains \cite{Textsummarization_using_LLM, Explore_Limits_ChatGPT_for_Text_Summarization}. They have also demonstrated effectiveness in supporting various facets of organizational decision-making, such as gathering information and generating potential options to consider \cite{goecks2024coagptgenerativepretrainedtransformers, LLM_present_new_questions_for_decisionsupport2024102811}.

\begin{figure}
    \centering
    \includegraphics[width=0.48\textwidth]{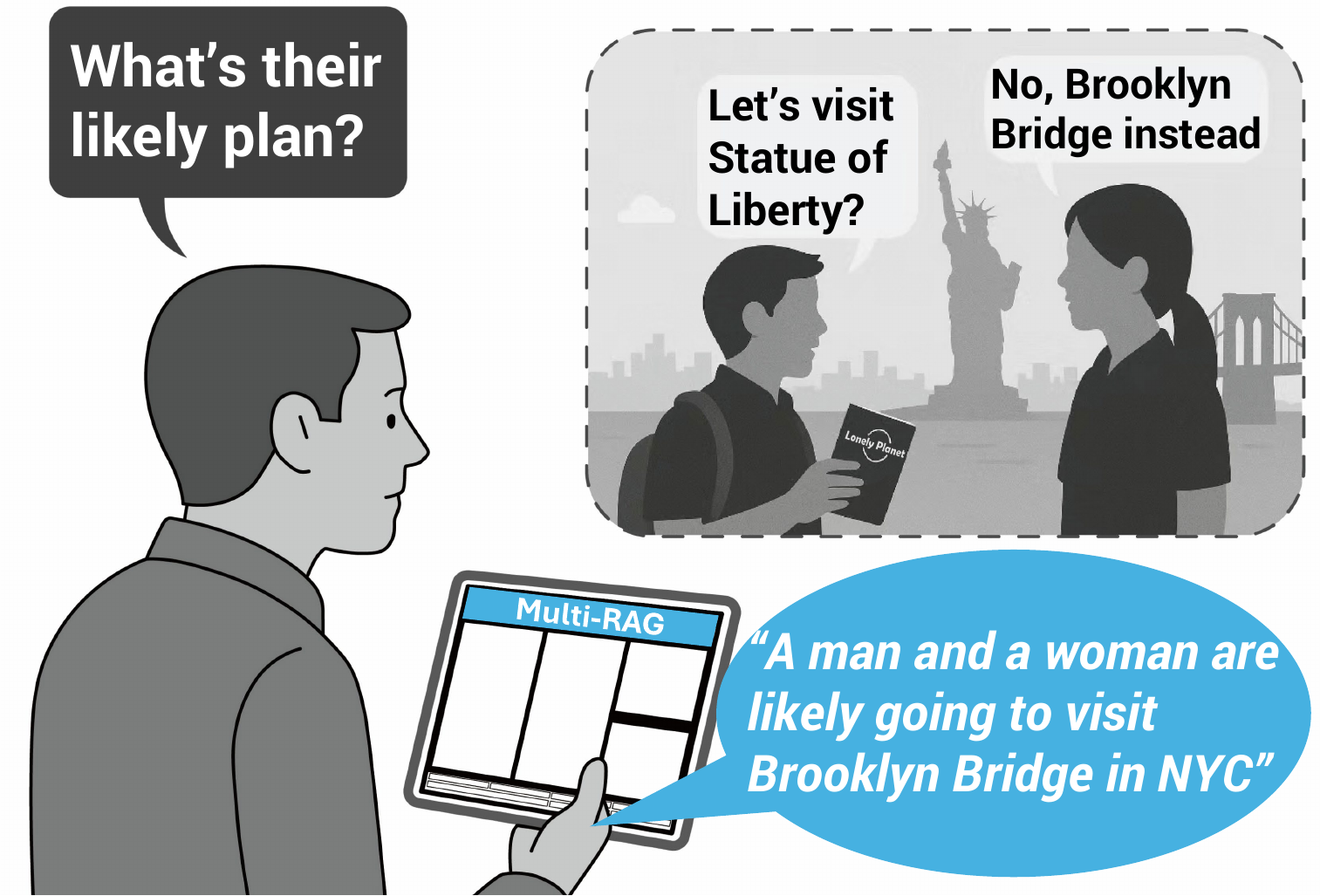}
    \caption{An example interaction scenario with the Multi-RAG system: Based on an open-ended user query, the system synthesizes visual, audio, and textual inputs to infer the answer, demonstrating its ability for practical understanding and decision support.}
    \label{fig:application_example}
\end{figure}

However, despite their growing popularity and capabilities, LLMs still face notable limitations in technical reliability and practical applicability. One major constraint is hallucinations, as LLMs are known to occasionally generate plausible-sounding but incorrect information \cite{survey_of_LLM}. Additionally, LLMs often lack the latest information because their training corpus is typically outdated by several months or years \cite{Hallucination_survey}. Therefore, we propose a human-centered system that helps the agent understand the world and increase its situational awareness across multi-modal scenarios, ultimately benefiting humans.

We developed an integrated software-hardware system that leverages Retrieval-Augmented Generation (RAG) to synergistically combine the intrinsic knowledge of LLMs with the vast and continuously evolving information from external databases. In addition, the system incorporates advanced perception capabilities—such as automatic speech recognition (ASR) and vision-language models (VLMs)—to enable three key functions: (1) perceiving and interpreting multimodal external inputs, (2) continuously acquiring and updating knowledge from dynamic real-world sources, and (3) enhancing human cognition by providing timely, context-aware support across various contexts (see Figure~\ref{fig:application_example}).

To demonstrate the system’s effectiveness in supporting human cognition and decision making, we focus our evaluation on complex, real-world scenarios that require the integration of multimodal information streams. Specifically, we employ an open-source video understanding benchmark as a representative task, as video-based environments naturally embody high information density and require both perceptual and reasoning capabilities—core aspects of effective decision support. Experimental results show that our system effectively interprets complex visual scenes and maintains high performance even when provided with limited visual input, highlighting its efficiency and robustness in resource-constrained scenarios.

\section{Related Works}
\label{sec:related_works}


Our work is closely related to, and draws inspiration from, prior research on memory augmentation systems, multi-modal retrieval, and intelligent assistants.

\subsection{Memory Augmentation Systems}
Memory augmentation systems have been an active area of research since Vannevar Bush's conception of the Memex in 1945 \cite{bush1945as}. Building on this foundation, wearable memory augmentation devices have gained increasing attention since the 1990s, evolving from early concepts such as "memory prosthesis" \cite{Human_Memory_Prosthesis}. Previous work in memory augmentation has placed significant emphasis on the audio modality. Studies have explored personal audio memory aids that support keyword-based search, such as Vemuri et al.'s system \cite{Audio-Based_Personal_Memory_Aid}. Additionally, researchers have investigated audio life-logging using wearable microphones and developed various smartphone-based browsing methods and real-time discrete and minimally disruptive querying mechanisms to access the recorded content \cite{Audio_Life-log_Data_Using_AcousticandLocation, memoro}.  With the development of multimedia technology, we hope the future memory augmentation media is not limited in the audio, therefore, we present a multimodal information retrieval method that enables users to retrieve information from captured video and other file types. 

\subsection{Multimodal RAG Paradigms}

Multimodal RAG systems can be implemented through a variety of distinct paradigms, each with unique approaches to handling multimodal embedding, retrieval, and generation \cite{abootorabi2025askmodalitycomprehensivesurvey}. 

\subsubsection{Cross-Modal}
Models like CLIP \cite{learning_transferable_visual_models}, ALBEF \cite{li2021alignfusevisionlanguage_ALBEF} enable both images and text to be encoded into a shared vector space, facilitating multimodal retrieval within existing text-centric  frameworks. This approach primarily requires replacing the unimodal embedding model with a multimodal encoder and using a multimodal LLM for generation, allowing unified question answering across modalities.

\subsubsection{Separate Modality Pipelines}


This approach treats each modality independently, designing dedicated retrieval and generation pipelines for each. Modality-specific encoders are used to embed queries and retrieve relevant content within the same modality, while generation relies on separate models tailored to each type of data. A re-ranking component can further optimize selection by identifying the most relevant results across different modality-specific retrieval stores \cite{mortaheb2025rerankingcontextmultimodalretrieval}.

\subsubsection{Single Primary Modality}
The single modality paradigm grounds all multimodal information into the embedding space of a primary modality, which is selected based on the focus of the application. Both retrieval and generation are then primarily performed using models specialized for the primary modality, allowing the system to standardize processing and leverage the strengths of typically text-based pipelines.

\subsection{Existing Assistants}
MerryQurry is an LLM-powered educational assistant that provides personalized, source-grounded support for students and educators, while addressing concerns around academic integrity and AI overreliance \cite{MerryQuery}. Expanding LLMs to domain-specific tasks, Kallakurik et al. \cite{kallakurik2025enabling} propose a medical AI assistant for edge devices that uses input-driven saliency to prune irrelevant neurons, enabling efficient deployment under hardware constraints. Moving beyond language-only inputs, STREAMMIND introduces a video-language framework for real-time streaming comprehension and dialogue, allowing AI systems to proactively respond to video content using context-aware natural language generation \cite{ding2025streammindunlockingframerate}.

\section{SYSTEM DESIGN}

\begin{figure*}[h]
    \centering
    \includegraphics[width=\textwidth]{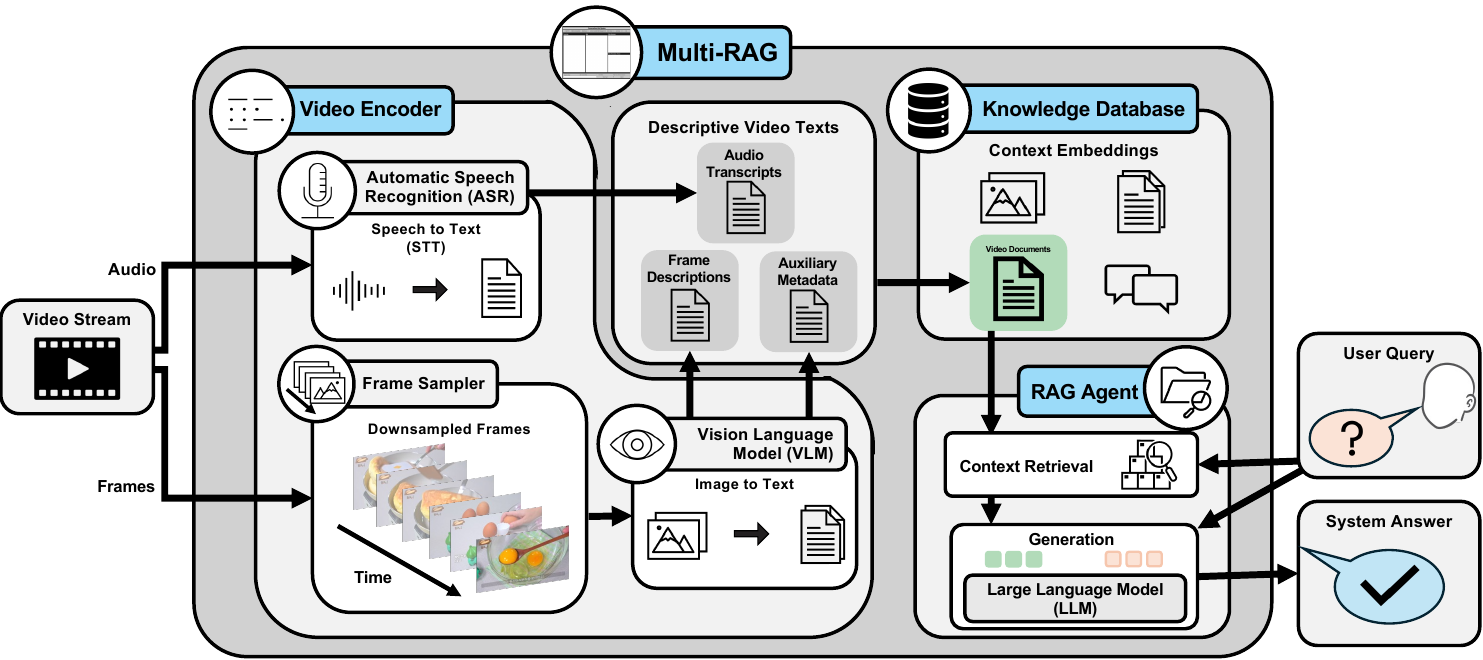}
    \caption{The framework of our Multi-RAG pipeline contains three key modules. In the video encoder, the VLM, receives input from the frame sampler, and the ASR processes visual and audio streams, respectively, converting multimodal external inputs into unified textual descriptions and embeddings. Next, these are stored in the knowledge vector database, where a retrieval-augmented generation (RAG) agent selects relevant information in response to the user query. The LLM then synthesizes the retrieved content to generate the final system response.}
    \label{fig:system_diagram}
\end{figure*}

Building on the single primary modality paradigm discussed in Section \ref{sec:related_works}, our system adopts this approach to build the video encoder, knowledge database, and RAG pipeline shown in Figure \ref{fig:system_diagram}. Specifically, it integrates multimodal information from video, audio, and documents, by converting all inputs into unified textual representations \cite{multi-modality-rag-comprehensivesurvey}. 
Visual and audio data are processed into descriptive text and auxiliary metadata, which are indexed in a vector database, while the original files are archived for potential future reference. At inference, user queries are matched to relevant content through semantic retrieval, and a large language model generates the final response based on both the query and retrieved information. In the following sections, we detail the video encoder framework, the construction of the knowledge database, and the RAG pipeline.

\subsection{Video Encoder} \label{subsec:video_encoder}

The processing of video data is bifurcated into two parallel streams: image processing and audio processing. For the video component, a sampling module is employed to select requisite frames. We have developed two distinct sets of sampling strategies tailored for two specific scenarios: educational environments and general-purpose applications. Specifically, within the context of classroom settings, the visual content predominantly consists of presentation slides. Lecturers typically expound upon a single visual frame for an extended duration, rendering a high frequency of input unnecessary. Consequently, for the visual stream, we have implemented a pre-processing stage designed to filter the input frames. Only upon detection of a substantive change in the visual content is the image subsequently fed into the VLM. Mean Square Error is used to measure the discrepancy between frame $A$ and frame $B$ of size $M\times N$ with $C$ channels: 

\begin{equation}
\mathrm{MSE}(A,B) = \frac{1}{M\times N\times C}\sum_{k=0}^{C-1}\sum_{i=0}^{M-1}\sum_{j=0}^{N-1}\bigl(A_{ijk} - B_{ijk}\bigr)^{2}
\end{equation}

To determine whether a newly sampled frame contains sufficient visual change for further processing, we compare its MSE against a cutoff value derived from sample videos of a recorded lecture series.
\sloppy
In general-purpose scenarios characterized by high-frequency transformations of visual content, the utility derived from input-filtering in the pre-processing stage is notably diminished. Consequently, for tasks of this nature, a uniform sampling methodology is employed. For the given video, the frame sampler samples $n$ frames at a constant rate. These frames $\{F_0, F_1, ...F_n \}$ are fed into the Vision Language Model to generate the text descriptions $\{D_0 , D_1 , ... D_n \}$ with timestamps. In addition to descriptions of the depicted content, auxiliary metadata are also generated, which include summaries and preliminary analyses \cite{luo2024videoragvisuallyalignedretrievalaugmentedlong}. Temporal summaries are generated at fixed intervals or upon slide transitions using prompts that ensure preservation of global context and mitigate overemphasis on individual frames. Preliminary analysis explains the video’s principal thematic focus, its predominant stylistic characteristics, and salient scene transitions. These auxiliary outputs augment frame-level descriptions by providing a macroscopic account of content progression alongside an initial interpretive framework of the video’s subject matter and structure.

Audio also encompasses a considerable volume of information. For corresponding video frames, audio recognition is implemented using OpenAI's Whisper Model—a general-purpose speech recognition model that supports automatic multilingual speech recognition. Audio information is converted into text using this model and appended to the frame's image text descriptions and auxiliary texts.

\subsection{Knowledge Database}




The choice of source file formats and the granularity of retrieval units profoundly influence the quality of subsequent generative outputs. Markdown has emerged as a preferred format for large language models due to several advantages: (1) it imposes a clear hierarchical structure through standardized syntax for headings, lists, and other organizational elements; (2) it supports typographic emphases—such as bolding, italics, and code spans—that facilitate nuanced semantic distinctions; and (3) it remains inherently human-readable, thereby enabling seamless collaboration between automated systems and human reviewers. As such, the text descriptions generated in Section \ref{subsec:video_encoder} are formatted as Markdown files. To build the knowledge database, the content of these documents is chunked and embedded, or encoded as vectors, for storage within a unified vector space.

\subsubsection{Chunking}
Chunking of documents before embedding ensures manageable input sizes to embedding models, preserves content, and reduces computational load, thus improving embedding quality and efficiency \cite{devlin2019bertpretrainingdeepbidirectional, vaswani2023attentionneed, mikolov2013efficientestimationwordrepresentations}.
 To chunk contextual documents, files are first loaded via format-appropriate parsers. During ingestion, each document is assigned a unique identifier, and basic metadata is recorded alongside the text chunk. To ensure that individual textual segments operate within the contextual window of the LLM while preserving meaningful semantic units and temporal information, documents are tokenized and partitioned into fixed-length chunks with an overlapping ratio. This approach guarantees that concepts spanning chunk boundaries are not lost.

\subsubsection{Embedding}
We utilized LangChain\cite{chase_langchain_2022} to generate embeddings for the contextual documents, enabling seamless integration of the LLM into the retrieval pipeline described in Section \ref{subsec:retrieval_agent} \cite{gao2024retrievalaugmentedgenerationlargelanguage_Survey}. Each text chunk is passed through an embedding model (OpenAIEmbeddings\cite{openai_embeddings}), and the output dense vector captures the chunk’s semantic content in a continuous space. The resulting vectors are stored in a vector database (Chroma\cite{chroma_2023}). 
\subsection{RAG Agent}\label{subsec:retrieval_agent}
The retrieval pipeline takes inputs from the knowledge database and user query to conduct a similarity search and generate context-aware system responses.

\subsubsection{Retrieval} Our system retrieves the top $k$ chunks most relevant to the user query based on semantic similarity. Like the document chunks, queries undergo vectorization. Retrieval is achieved by computing the cosine similarity between the embedding of the query $\mathrm{Q}$ and the document chunk $\mathrm{C}$.

\begin{equation}
\begin{aligned}
    \{\mathrm{C}_{i_1}, \ldots, \mathrm{C}_{i_k}\} 
    =\ \operatorname{arg\,top}_k\, \Big\{\, 
    &\mathrm{Cosine\_similarity}(\mathbf{Q},\, \mathbf{C}_j) \mid \\
    &j = 1, \ldots, N\, \Big\}
\end{aligned}
\end{equation}


\subsubsection{Generation} The original input query and the retrieved chunks are fed into LLM to generate the final system answer. In our system, we employ two distinct prompt types to guide the model’s response generation. The first prompt type is designed to encourage open-ended reasoning and creative inference, allowing the model greater flexibility in generating explanations and predictions. The second prompt type, by contrast, places a stronger emphasis on factual grounding and concise answers, promoting a more balanced trade-off between creativity and accuracy. 



\section{Experimentation}

To rigorously assess the effectiveness of our system in supporting multimodal reasoning, we conducted experiments using standardized video understanding benchmarks and systematic parameter analyses.

\subsection{Supporting Datasets}

The MMBench-Video benchmark \cite{mm_video_bench} is designed to comprehensively evaluate the video understanding capabilities of LVLMs. It features long-form YouTube videos and uses open-ended questions to reflect real-world application scenarios. The dataset includes \textbf{609} video clips across 16 major categories such as news, science, and finance. Video durations range from 30 seconds to 6 minutes, with a total of 1998 question-answer(QA) pairs.

\begin{figure}
    \centering
    \includegraphics[width=0.5\textwidth]{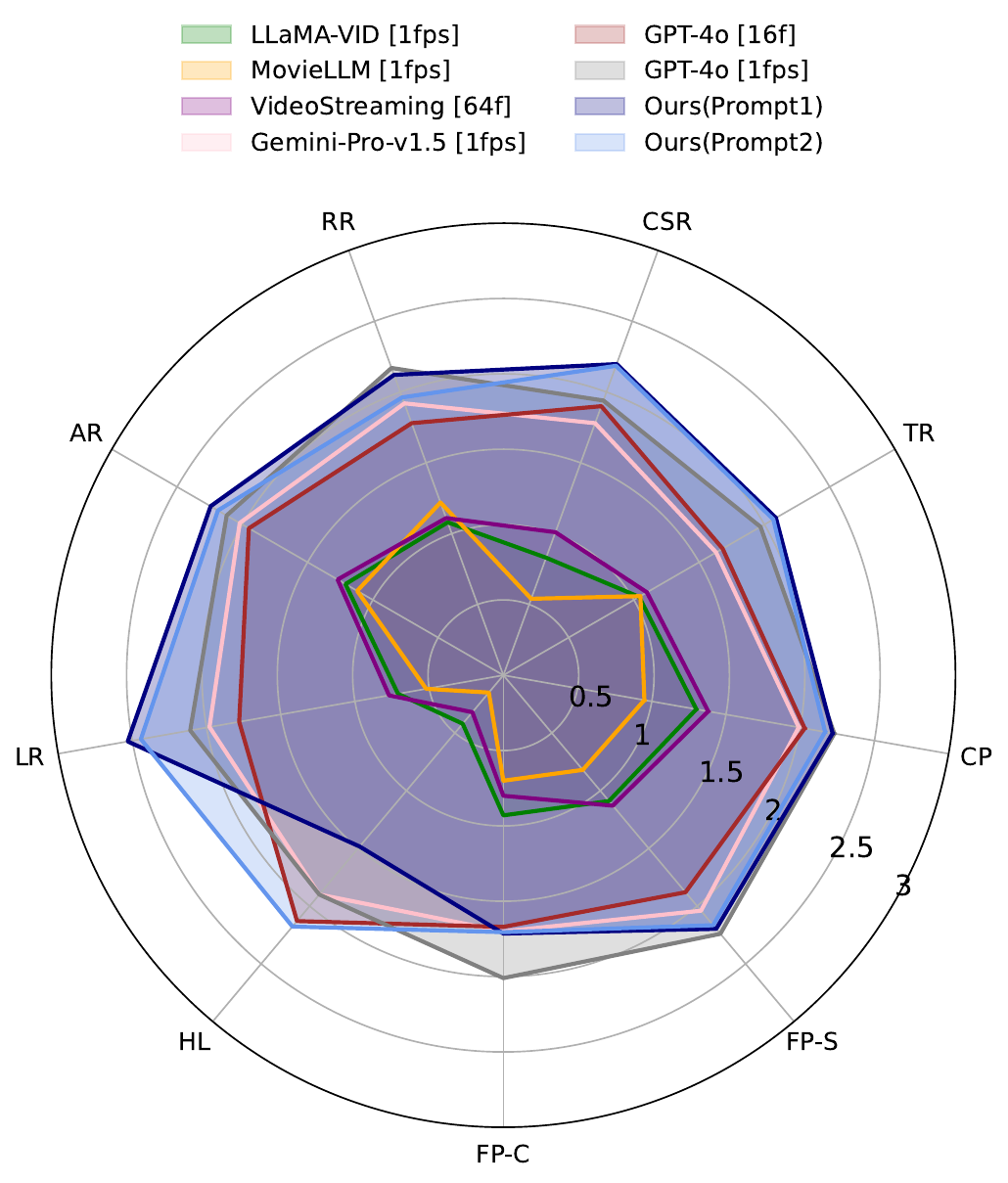}
    \caption{Comparison of mainstream Video-LLMs, LVLMs, and our system on MMBench-Video. This graph illustrates performance in the Level 2 capabilities from the QA setting in the MMBench-Video benchmark. Those are Coarse Perception (CP), Fine-grained Perception with Single-Instance (FP-S), Fine-grained Perception with Cross-Instance (FP-C), Hallucination (HL), Logic Reasoning (LR), Attribute Reasoning (AR), Relation Reasoning (RR), Common Sense Reasoning (CSR), and Temporal Reasoning (TR).}
    \label{fig:ridar_results}
\end{figure}

\begin{table*}
\centering
\caption{Comparison of Video-LLMs on Perception and Reasoning Tasks}
\label{tab:video_llm_results}
\setlength{\tabcolsep}{6pt}
\begin{tabular}{l|c|cccc|c| ccccc|c}
\hline
\multirow{2}{*}{\textbf{Model}} & \textbf{Overall} 
& \multicolumn{5}{c|}{\textbf{Perception}} 
& \multicolumn{6}{c}{\textbf{Reasoning}} \\
\cline{3-13}
& \textbf{Mean}& \textbf{CP} & \textbf{FP-S} & \textbf{FP-C} & \textbf{HL} & \textbf{Mean} 
  & \textbf{LR} & \textbf{AR} & \textbf{RR} & \textbf{CSR} & \textbf{TR} & \textbf{Mean} \\
\hline
\multicolumn{13}{c}{Open-source Video-LLMs} \\
\hline
LLaMA-VID-[1fps]    & 1.08 & 1.30 & 1.09 & 0.93 & 0.42 & 0.94 & 0.71 & 1.21 & 1.08 & 0.83 & 1.04 & 1.02 \\
MovieLLM-[1fps]     & 0.87 & 0.95 & 0.82 & 0.70 & 0.15 & 0.65 & 0.52 & 1.12 & 1.22 & 0.54 & 1.05 & 0.97 \\
VideoStreaming-[64f+]  & 1.12 & 1.38 & 1.13 & 0.8 & 0.32 & 1.13 & 0.77 & 1.27 & 1.11 & 1.01 & 1.10 & 1.09 \\
\hline
\multicolumn{13}{c}{Open-source LVLMs} \\
\hline
InternVL2-26B-[16f]   & 1.41 & 1.56 & 1.48 & 1.23 & 0.52 & 1.42 & 1.06 & 1.61 & 1.45 & 1.38 & 1.23 & 1.35 \\
VILA1.5-40B-[14f] & 1.61 & 1.78 & 1.72 & 1.35 & 0.47 & 1.63 & 1.12 & 1.78 & 1.61 & 1.48 & 1.45 & 1.52 \\
\hline
\multicolumn{13}{c}{Proprietary LVLMs for Images} \\
\hline
Gemini-Pro-v1.5-[1fps] & 1.94 & 1.99 & 2.04 & 1.70 & 1.90 & 1.98 & 1.98 & 2.02 & 1.92 & 1.78 & 1.63 & 1.86 \\
GPT-4o-[1fps]          & 2.15 & 2.23 & 2.24 & 2.01 & 1.90 & 2.19 & 2.11 & 2.12 & 2.17 & 1.94 & 1.97 & 2.08 \\
\hline
\multicolumn{13}{c}{Ours} \\
\hline
\textbf{Multi-RAG, 0.5fps, prompt1} & \textbf{2.14} & 2.20 & 2.14 & 1.71 & 1.48 & \textbf{2.08} & \textbf{2.53} & \textbf{2.24} & 2.12 & \textbf{2.20} & \textbf{2.09} & \textbf{2.21} \\
\textbf{Multi-RAG, 0.5fps, prompt2} & \textbf{2.12} & 2.17 & 2.17 & 1.71 & 2.18 & \textbf{2.11} & 2.44 & 2.19 & 1.96 & 2.19 & 2.07 & \textbf{2.14} \\
\hline
\end{tabular}
\end{table*}

The QA setting in the MMBench-Video benchmark is structured around a three-level hierarchical capability taxonomy. At the highest level (Level 1), capabilities are divided into two domains: Perception and Reasoning. These are further subdivided into nine Level 2 capabilities: Coarse Perception (CP), Fine-grained Perception with Single-Instance (FP-S), Fine-grained Perception with Cross-Instance (FP-C), Hallucination (HL), Logic Reasoning (LR), Attribute Reasoning (AR), Relation Reasoning (RR), Common Sense Reasoning (CSR), and Temporal Reasoning (TR). Level 3 capabilities consist of 26 fine-grained categories, capturing a broad spectrum of cognitive skills relevant to video understanding.

\subsection{Evaluation Procedures}

In evaluating the performance of our video understanding system, we also assess its operational efficiency. This is investigated by systematically reducing the input frame rate to determine the necessity of high-frequency sampling. Concurrently, comprehensive ablation studies are conducted to elucidate the influence of other critical parameters, including the integration of audio information, the number of retrieved contextual chunks, and variations in question generation prompt design.

The experimental procedure commenced with the extraction of video frames at a rate of 1 frame per second (fps); corresponding textual descriptions, timestamped and chronologically ordered, were then recorded. An initial evaluation relevant to the benchmark was conducted using this 1 fps dataset.

Subsequently, the frame rate was effectively halved to 0.5 fps by removing alternate frames, and a second evaluation was performed. Finally, an Automatic Speech Recognition (ASR) module was employed to transcribe the video's audio content. This transcribed audio information was then integrated into the respective datasets for both the 1 fps and 0.5 fps conditions. Subsequently, as detailed in Section \ref{subsec:video_encoder}, documents were augmented with auxiliary contextual information to facilitate a further evaluation of the results. To reduce operational costs and latency, GPT-4.1 mini was chosen over GPT-4o, despite offering no additional improvements in model intelligence. Aligned with the evaluation methodology established in the MMBench-video, we employed GPT-4 to assess the discrepancy between the model's generated answer and ground truth answers.




\section{Results}


We present the best configuration of our system in Table~\ref{tab:video_llm_results} and compare it with the strongest baseline, GPT-4o at 1 fps input. Notably, our system achieves comparable or even superior performance while operating at only half the input frame rate (0.5fps). With prompt1, our system attains an overall mean of 2.14, nearly matching GPT-4o, and even surpasses GPT-4o in reasoning mean (2.21 vs. 2.08). Similarly, with prompt2, the overall mean remains high at 2.12, and the system demonstrates a balanced improvement in both perception and reasoning, with particularly strong performance in perception hallucination when using prompt2. Figure~\ref{fig:ridar_results} intuitively shows the capabilities of each model against our system in the different tasks. 
\subsection{Ablation Studies}
\begin{table}
\centering
\caption{Results of Ablation Study.}
\label{tab:abalation}
\setlength{\tabcolsep}{0.8pt} 
\begin{tabular}{l|c|c|c}
\hline
\textbf{Methods} & \makecell{\textbf{Perception}\\\textbf{Mean}} & \makecell{\textbf{Reasoning}\\\textbf{Mean}} & \makecell{\textbf{Overall}\\\textbf{Mean}} \\
\hline
\textit{Multi-RAG} &  2.12 & 2.14 & 2.12 \\
\hline
\textit{Multi-RAG} \textit{w/o} Audio & 2.03 & 2.02 & 2.03\\
\hline
\textit{Multi-RAG} \textit{w/o} Auxiliary metadata & 2.09 & 2.12 & 2.10\\
\hline
\textit{Multi-RAG} \textit{w/o} Audio \& Auxiliary metadata & 2.02
  & 1.98 & 2.01\\
\hline
\end{tabular}
\end{table}

\begin{table}
\centering
\caption{Results of System with Different Retrieved Chunks.}
\label{tab:param_k}
\setlength{\tabcolsep}{10pt} 
\begin{tabular}{c|c|c|c|c}
\hline
\textbf{Top-k} & \textbf{Methods} & \makecell{\textbf{Perception}\\\textbf{Mean}} & \makecell{\textbf{Reasoning}\\\textbf{Mean}} & \makecell{\textbf{Overall}\\\textbf{Mean}} \\
\hline
Top-1 & \textit{Multi-RAG} & 1.93 & 1.90 & 1.92 \\
\hline
Top-3 & \textit{Multi-RAG} & 2.03  & 2.06  &  2.04\\
\hline
Top-5 & \textit{Multi-RAG} & 2.12 & 2.14 & 2.12 \\
\hline 
Top-7 & \textit{Multi-RAG} & 2.08 & 2.15 & 2.10 \\
\hline
\end{tabular}
\end{table}
Overall, these results highlight the efficiency and effectiveness of our proposed system. It matches or exceeds the state-of-the-art proprietary models on several tasks, particularly in reasoning, while demonstrating robust performance even at reduced frame rates. This underscores the benefits of prompt design and multimodal fusion in achieving high performance across a variety of video understanding tasks.


We performed ablation studies to quantify the contributions of two key components—audio and textual auxiliary metadata—to the overall system performance, using prompt2 as the baseline configuration. As shown in Table~\ref{tab:abalation}, the results highlight the effectiveness of the evaluated components. Audio information contributes substantially to overall performance, while auxiliary text offers a smaller but still positive impact. Notably, the system exhibits the lowest performance when both components are removed, underscoring their complementary roles in the system.

\subsection{Parameter Analysis}

As shown in Table~\ref{tab:param_k}, increasing the number of retrieved chunks (top-k) generally leads to improved system performance. Both perception mean and reasoning mean metrics increase as top-k rises from 1 to 5, reaching optimal values at top-5. However, further increasing top-k to 7 yields only marginal changes, indicating diminishing returns. These results, also obtained with the prompt 2, suggest that retrieving a moderate number of relevant chunks (such as top-5) strikes a good balance between information coverage and retrieval efficiency.



\section{Conclusion}

As robots increasingly integrate into the fabric of human society, enabling them to learn continually from human interaction is both a technical imperative and a societal opportunity. Our Multi-RAG system demonstrates how multi-modal retrieval-augmented generation can serve as a scalable, adaptive foundation for robotic systems that support human cognition in dynamic, information-rich environments. By grounding perception, audio, and image in real-time, human-centered inputs, this work bridges the gap between static AI models and socially aware, learning-enabled agents. 

Moving forward, we plan to enhance the system's user interface to further optimize human cognitive support, ensuring more intuitive and effective interactions for users in real-world settings. Additionally, we will explore adaptive retrieval strategies that dynamically allocate computational resources based on real-time context, aiming to minimize system overhead while maintaining robustness and responsiveness across complex and evolving scenarios.

We hope this work fosters deeper interdisciplinary discussions and inspires novel frameworks that push the frontier of continual robot learning, from memory augmentation to human-norm acquisition, ultimately empowering robots to become more capable, collaborative, and socially intelligent partners in our shared future.

\section*{Acknowledgments}
This work has been partially supported by the U.S. Army DEVCOM Army
Research Laboratory Cooperative Agreement No. W911NF2120211. We also thank Junsong Lin for providing valuable assistance with the figures in this paper.


\bibliographystyle{plainnat}
\bibliography{references}

\newpage
\section*{Appendix}

\subsection{Prompts Adopted in the Multi-RAG System}

\noindent\textbf{System Prompts for the RAG Pipeline:}\\
\textbf{Prompt Type 1}
\begin{lstlisting}
You are an assistant answering questions based primarily on the provided context.
Instructions:
1. Answer concisely and directly within one sentence.
2. Analyze the 'Context' provided below.
3. Synthesize information from the 'Context' to answer the 'User's Question'.
4. Even if the information is insufficient, guess the most possible answer. 
   Minimize negative responses and stimulate your imagination!
5. Base your answer mainly on the 'Context'. Avoid introducing external knowledge.
6. Combine information from different parts of the context if needed.
7. Do not use phrases like "According to the context...".
Context from Previous Conversation:
{context}
User's Question:
{question}
Response:
\end{lstlisting} 

\noindent\textbf{Prompt Type 2}
\begin{lstlisting}
You are an assistant answering questions based primarily on the provided context.
Instructions:
1. Answer concisely and directly within one sentence.
2. Analyze the 'Context' provided below.
3. Synthesize information from the 'Context' to answer the 'User's Question'.
4. If the context does not clearly contain the answer, you may offer the
   single most plausible guess -- but you MUST:
   - Prefix with "Speculative --"
   - Keep the guess concise (<=1 sentence).
   - Do NOT present speculation as fact.
   - If no reasonable guess exists, reply "Unknown."
   - Avoid empty phrases like "not possible to determine" unless "Unknown" is the only honest response.
   This lets you use imagination while making it clear what is evidence-based.
5. Base your answer mainly on the 'Context'. Avoid introducing external knowledge.
6. Combine information from different parts of the context if needed.
7. Do not use phrases like "According to the context...".
Context from Previous Conversation:
{context}
User's Question:
{question}
Response:
\end{lstlisting}

\noindent\textbf{System Prompt for Frame Description and Auxiliary Text}
\begin{lstlisting}
Describe this video frame concisely and capture the key information:
You are an AI assistant skilled at analyzing text descriptions of video content. 
Your goal is to synthesize this analysis into a single, coherent paragraph summarizing key aspects.
Analyze the following text describing video content. Then, write one single paragraph that weaves together these elements:
Topic: What is the main subject matter?
Emotion: What is the dominant mood or feeling?
Scene: Briefly describe the key visual setting or elements mentioned in the text.
Style: What is the likely style of the video?
\end{lstlisting}

\noindent\textbf{System Prompt for Evaluation}
\begin{lstlisting}
As an AI assistant, your task is to evaluate a candidate answer in comparison to a given correct answer.
The question, groundtruth answer, and candidate answer will be provided.
Your assessment should range from 0 to 3, based solely on semantic similarity:
- 0: No similarity (entirely incorrect)
- 1: Low similarity (largely incorrect)
- 2: High similarity (largely correct)
- 3: Complete similarity (entirely correct)
Your response should be a single integer from 0, 1, 2, or 3.

Question: [QUESTION]
Groundtruth answer: [ANNOTATED ANSWER]
Candidate answer: [CANDIDATE ANSWER]
Your response:
\end{lstlisting}

\begin{figure}[htbp]
    \includegraphics[width=0.48\textwidth]{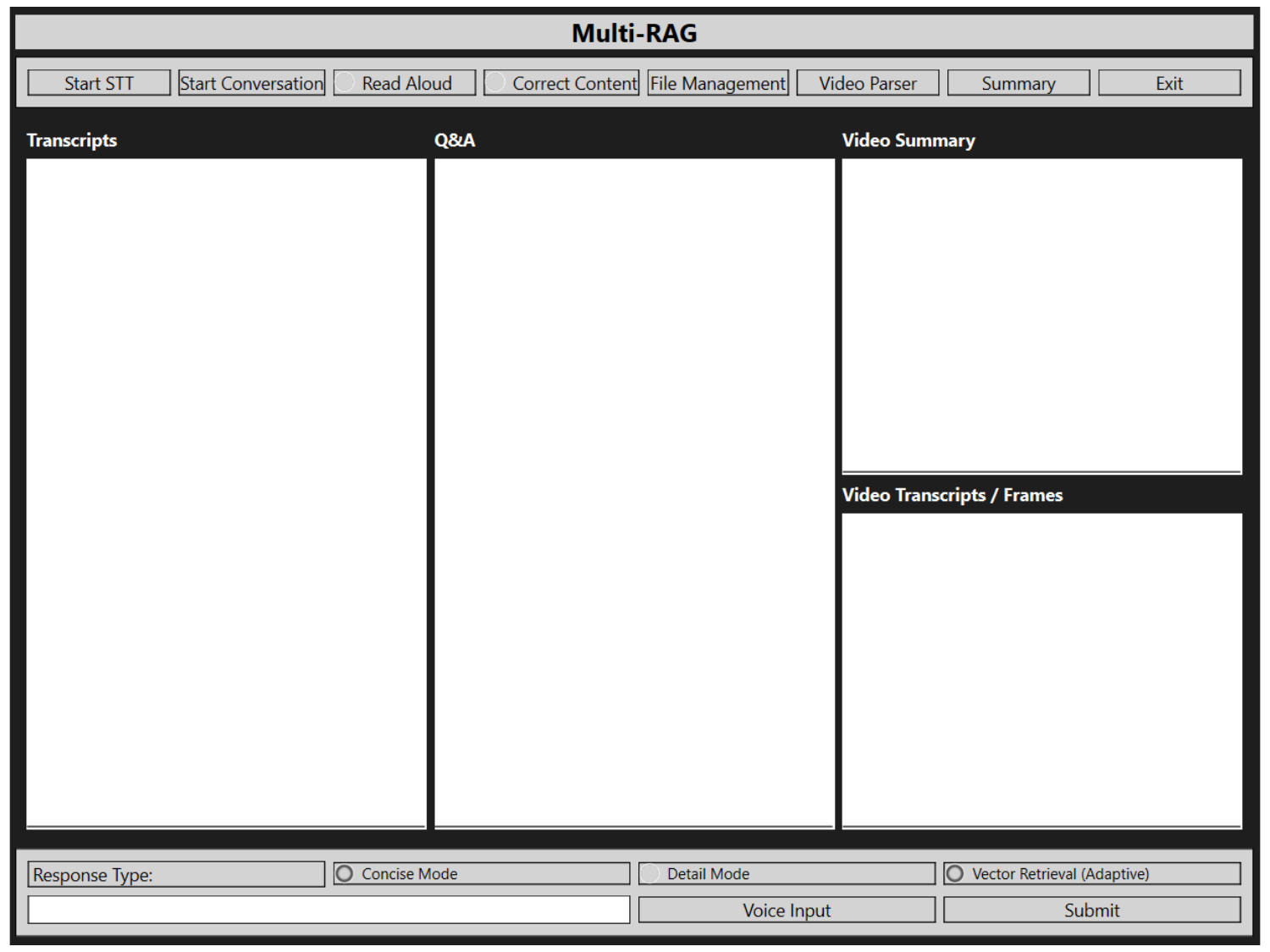}
    \caption{The Multi-RAG user interface is organized into three main columns. The left column serves as the dialogue section, displaying transcripts of ongoing conversations. The central column is dedicated to the Q\&A area, where the system presents answers to user queries. The right column focuses on video content and is divided into two parts: the upper section provides rolling or overall video summaries, while the lower section displays detailed video descriptions, including selected frames and corresponding audio transcripts.}
    \label{fig:GUI}
\end{figure}

\subsection{Experimental User interface}

We have implemented a graphical user interface to facilitate further experimentation. The top section features a function bar, providing controls for initiating or terminating speech recognition, real-time error correction, engaging in dialogue with the system, file management, video parsing, and summary requests. The main display area is divided into three columns: the left column serves as the dialogue panel, displaying transcripts of ongoing interactions; the central column is dedicated to the Q\&A area, where system responses to user queries are presented; and the right column is devoted to video content, with the upper section offering dynamic or overall video summaries and the lower section presenting detailed video descriptions, including representative frames and associated audio transcripts. The bottom section of the interface allows users to select the response type and input questions, supporting both voice and text modalities.

\subsection{Cost}

The total cost for API call of this experiment is approximately \$482.  
\end{document}